\documentclass{article}
\usepackage{float}
\usepackage{graphicx}
    \PassOptionsToPackage{numbers, compress}{natbib}

 \usepackage[dblblindworkshop, final]{neurips_2026}
\workshoptitle{AI4Science}

\usepackage[utf8]{inputenc} 
\usepackage[T1]{fontenc}    
\usepackage{hyperref}       
\usepackage{url}            
\usepackage{booktabs}       
\usepackage{amsfonts}       
\usepackage{nicefrac}       
\usepackage{microtype}      
\usepackage{xcolor}         
\usepackage{amsmath}
\usepackage{amssymb}
\usepackage{multicol}
\usepackage{multirow}
\usepackage{tabularx}
\usepackage{longtable}
\usepackage{array}
\usepackage[most]{tcolorbox}
\usepackage{enumitem}
\usepackage[most]{tcolorbox}
\usepackage{xcolor}

\definecolor{promptheader}{RGB}{95,95,95}
\definecolor{promptbody}{RGB}{248,248,248}
\definecolor{promptborder}{RGB}{110,110,110}

\newtcolorbox{promptbox}[1]{
  enhanced,
  breakable,
  width=\linewidth,
  colback=promptbody,
  colframe=promptborder,
  boxrule=0.7pt,
  arc=1.8mm,
  outer arc=1.8mm,
  left=4mm,
  right=4mm,
  top=4mm,
  bottom=3mm,
  title={#1},
  coltitle=white,
  fonttitle=\bfseries\small,
  attach boxed title to top left={xshift=3mm,yshift=-2mm},
  boxed title style={
    colback=promptheader,
    colframe=promptheader,
    boxrule=0pt,
    arc=1.3mm,
    outer arc=1.3mm,
    left=2.5mm,
    right=2.5mm,
    top=0.7mm,
    bottom=0.7mm
  },
  before skip=7pt,
  after skip=7pt
}
\usepackage[most]{tcolorbox}

\definecolor{successheader}{RGB}{88,125,102}
\definecolor{failureheader}{RGB}{145,93,93}
\definecolor{examplebody}{RGB}{249,249,249}
\definecolor{exampleborder}{RGB}{120,120,120}

\newtcolorbox{successbox}[1]{
  enhanced,
  breakable,
  width=\linewidth,
  colback=examplebody,
  colframe=exampleborder,
  boxrule=0.6pt,
  arc=1.8mm,
  outer arc=1.8mm,
  left=3.5mm,
  right=3.5mm,
  top=3.5mm,
  bottom=2.5mm,
  title={#1},
  coltitle=white,
  fonttitle=\bfseries\small,
  attach boxed title to top left={xshift=3mm,yshift=-1.8mm},
  boxed title style={
    colback=successheader,
    colframe=successheader,
    boxrule=0pt,
    arc=1.2mm,
    outer arc=1.2mm,
    left=2.3mm,
    right=2.3mm,
    top=0.6mm,
    bottom=0.6mm
  },
  before skip=7pt,
  after skip=7pt
}

\newtcolorbox{failurebox}[1]{
  enhanced,
  breakable,
  width=\linewidth,
  colback=examplebody,
  colframe=exampleborder,
  boxrule=0.6pt,
  arc=1.8mm,
  outer arc=1.8mm,
  left=3.5mm,
  right=3.5mm,
  top=3.5mm,
  bottom=2.5mm,
  title={#1},
  coltitle=white,
  fonttitle=\bfseries\small,
  attach boxed title to top left={xshift=3mm,yshift=-1.8mm},
  boxed title style={
    colback=failureheader,
    colframe=failureheader,
    boxrule=0pt,
    arc=1.2mm,
    outer arc=1.2mm,
    left=2.3mm,
    right=2.3mm,
    top=0.6mm,
    bottom=0.6mm
  },
  before skip=7pt,
  after skip=7pt
}
\usepackage{subcaption}
\title{\textsc{SciGraphSelect}: Auditable Scientific Data Selection through Knowledge–Reasoning Graphs}

\author{%
  MiHyeon Kim$^{*}$ \quad
  Ji-Eun Han$^{*}$ \quad
  Eunkyeong Lee$^{*}$\quad
  Jeehyun Lim \quad
  Hwijung Ryu$^{\dagger}$ \\
  KT Corporation \\
  \texttt{\{mihyeon.gim,ji-eun.han,ek.lee,jeehyun.lim,\}@kt.com} \\
  \\
  $^{*}$Equal contribution. \quad
  $^{\dagger}$Corresponding author.
}
\begin{document}

\maketitle

\begin{abstract}
Scientific data selection is typically driven by relevance or utility scores,
providing limited insight into why a document is retained. We introduce
\textsc{SciGraphSelect}, a reference-grounded framework for auditable
scientific data selection. SciGraphSelect organizes domain-specific knowledge,
reusable reasoning operations, and reference-attested relations into a typed
knowledge--reasoning graph, then scores candidate documents using concept
coverage, relation and path co-support, and directional alignment. To avoid
over-interpreting document embeddings, rich relational structure is preserved
on the reference side while candidate-side evidence is treated conservatively
without assuming relation entailment. By exposing the reference-grounded
evidence behind each selection, SciGraphSelect brings traceability to the
upstream construction of scientific training corpora and complements
downstream scientific verification.
\end{abstract}

\section{Introduction}

Scientific foundation models increasingly depend on large and heterogeneous corpora assembled from scientific papers, domain resources, and synthetic or curated data. Yet the usefulness of scientific data is not determined by topical relevance alone. Raw scientific literature can exhibit a substantial learnability gap \cite{qin2026scipedia}, document-level quality or domain labels can miss meaningful variation within scientific articles \cite{laiking2024biomedical,touchent2026biomed}, and domain corpora contain heterogeneous scientific entities, mechanisms, and relations \cite{seo2026biocollection,zhang2024chemllm}. Two documents can therefore be equally relevant to Biology or Chemistry while differing substantially in the knowledge they cover, the scientific relations they support, and the reasoning operations they instantiate.

This distinction becomes increasingly important as AI systems take on larger parts of the scientific workflow, including literature synthesis, hypothesis generation, experimentation, and evaluation \cite{lu2024aiscientist}. Trustworthy scientific AI consequently requires not only evaluating model outputs, but also understanding the evidence used to construct and train such systems. Recent work on scientific claim verification similarly emphasizes source-level accountability and evidence traceability \cite{tan2026scitrue}. We study a complementary upstream problem: \emph{can scientific training-data selection itself expose auditable evidence for why a document is retained?}

Most large-scale data-selection methods optimize distributional similarity, diversity, data mixture, or estimated training utility \cite{xie2023dsir,xie2023doremi,tirumala2023d4,yu2024mates}. These objectives improve pretraining efficiency, but generally reduce each candidate to a scalar notion of usefulness without representing the scientific structure responsible for that decision. MASS takes an important step toward structure-aware selection by deriving a skill graph from mathematical references \cite{li2025mass}. Scientific corpora, however, pose a different representational challenge. Domain-specific \emph{knowledge}---such as molecular mechanisms, biological processes, structures, reactions, and physical properties---must be distinguished from reusable \emph{reasoning operations}, such as calculation, comparison, mechanistic analysis, structure elucidation, and evidence interpretation. Moreover, scientific relations should remain traceable to reference evidence rather than being inferred solely from semantic proximity.

A naive solution would be to construct a rich scientific graph and treat every graph match in a candidate document as evidence that the corresponding relation is expressed. This is too strong when corpus-scale scoring relies on whole-document representations: similarity to two related concepts does not establish that the document entails their directed relation, nor does activation of a reasoning path reveal a latent chain of thought. The central challenge is therefore to exploit rich scientific structure \emph{without making stronger document-level claims than the available evidence supports}.

We introduce \textsc{SciGraphSelect}, a reference-grounded framework for auditable scientific data selection. From domain-matched Biology and Chemistry references, SciGraphSelect extracts domain-specific knowledge concepts \(K\), reusable reasoning operations \(R\), and reference-attested relations among them, forming a typed directed acyclic graph with \(K\!\rightarrow\!K\), \(K\!\rightarrow\!R\), and \(R\!\rightarrow\!R\) edges. Each graph element retains its reference provenance. Candidate documents are then scored using knowledge and reasoning coverage together with conservative edge and path co-support. Crucially, graph structure determines \emph{which combinations of evidence matter}, while document scoring does not assume that entail the corresponding reference relation
jointly supported nodes entail the corresponding reference relation. Directionality is incorporated separately through relative alignment between forward and reversed relation verbalizations.

We evaluate SciGraphSelect on Biology and Chemistry using the PeS2o~\citep{peS2o} scientific-paper pool and a heterogeneous hard-negative setting.
Across both domains, reference-grounded scoring improves target-domain
selection over embedding similarity, while component-wise analysis shows that
concept, relation, reasoning, and directional signals contribute in distinct
and domain-dependent ways.
Beyond ranking performance, these analyses reveal \emph{why} a document is
selected---whether through broad concept coverage, reference-consistent
relations, procedural reasoning, or directional compatibility with scientific
dependencies.
We further complement source-proxy evaluation with human judgments of
selection relevance and attribution support, allowing us to assess not only
which documents are preferred, but also whether the evidence used to justify
those selections is interpretable and faithful to the source text.
Our contributions are threefold:
\begin{itemize}
\item We formulate scientific data selection using
{reference-grounded knowledge--reasoning structure}, separating
domain-specific knowledge concepts from reusable reasoning operations while
retaining their reference provenance.

\item We propose a document-scoring method that combines
{knowledge/reasoning coverage, relation and path co-support, and
directional alignment}, while explicitly distinguishing these signals from
document-level relation entailment.

\item We evaluate the resulting selections across Biology and Chemistry,
showing how the different signals alter the selected corpus and complementing
source-proxy evaluation with {human judgments of document relevance
and attribution support}.
\end{itemize}

\section{Related Work}

\paragraph{Pretraining data selection and scientific curation.}
Large-scale pretraining has motivated data-selection methods based on distribution matching, embedding geometry, domain mixtures, and model-dependent utility. DSIR uses importance resampling \cite{xie2023dsir}, D4 combines semantic deduplication and diversification \cite{tirumala2023d4}, DoReMi optimizes domain-mixture weights \cite{xie2023doremi}, and MATES estimates data influence during pretraining \cite{yu2024mates}. MASS is the closest structural precedent, constructing a skill graph from mathematical references to guide data selection \cite{li2025mass}. Scientific-data work has instead emphasized domain-specific curation and learnability: biomedical selection has used journal-level quality signals \cite{laiking2024biomedical}, Biomed-Enriched introduces paragraph-level annotations \cite{touchent2026biomed}, and SciPedia addresses the learnability gap of raw scientific literature through cleaning and pedagogical augmentation \cite{qin2026scipedia}. Domain resources such as TheBioCollection \cite{seo2026biocollection} and ChemLLM \cite{zhang2024chemllm} further demonstrate the value of specialized scientific corpora. SciGraphSelect complements these approaches by representing not only whether a document is useful or domain-relevant, but also which knowledge, reasoning, and reference-grounded structural evidence supports its selection.

\paragraph{Scientific structure and evidence-grounded verification.}
Recent work has explored LLM-based construction of structured scientific knowledge. Oarga et al.\ generate scientific ontologies and knowledge graphs with open LLMs \cite{oarga2026scientifickg}, while SciGraph-LLM builds provenance-aware knowledge graphs grounded in scientific papers \cite{malashin2026scigraphllm}. SciGraphSelect uses such structure for a different purpose: rather than treating the graph as the final representation, it uses reference-attested knowledge and relations to guide corpus selection. Importantly, the reference graph defines which concept combinations are structurally meaningful, while candidate-document scoring remains conservative and does not interpret relation co-support as relation entailment. This also distinguishes our setting from scientific claim verification. Systems such as SciTrue link claims to explicit source evidence for accountable verification \cite{tan2026scitrue}; SciGraphSelect instead applies the same principle of evidence traceability to the upstream selection of scientific training data.

\section{\textsc{SciGraphSelect}}
\label{sec:method}

\begin{figure*}[t]
    \centering
    \includegraphics[width=\textwidth]{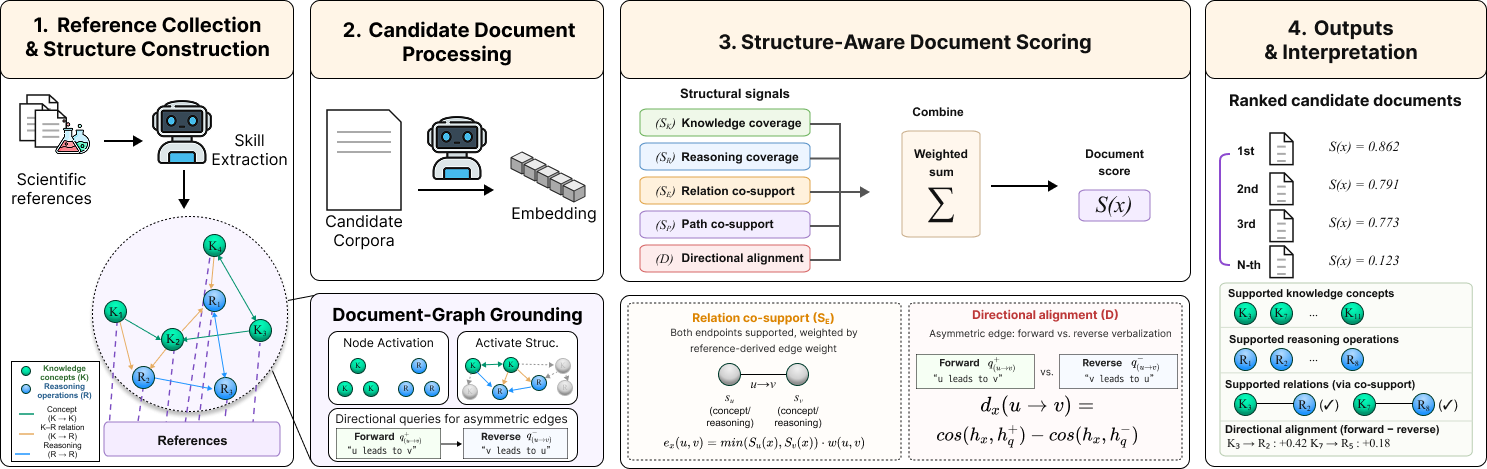}
    \caption{
\textbf{Overview of \textsc{SciGraphSelect}.}
Reference-derived knowledge--reasoning structure is matched against candidate
documents to compute structure-aware selection scores and expose the evidence
supporting each ranked document.
}
    \label{fig:scigraphselect_overview}
\end{figure*}

SciGraphSelect is a reference-grounded framework that structures scientific
knowledge and reasoning from domain references, then uses this structure to
select training data based on the knowledge and reasoning supported by each
document. Figure~\ref{fig:scigraphselect_overview} provides an overview of the
full pipeline, from reference-graph construction and candidate-document
processing to structure-aware scoring and interpretable document ranking.

\subsection{Reference-Grounded Concept--Reasoning DAG}
\label{sec:reference_dag}

Given scientific references, we extract two complementary node types.
A knowledge node $k\in\mathcal{K}$ represents domain-specific scientific
content, such as mechanisms, structures, reactions, properties, or laws,
whereas a reasoning node $r\in\mathcal{R}$ represents reusable operations
applied to that knowledge, such as calculation, comparison, mechanistic
analysis, structure elucidation, or evidence interpretation.
The extraction procedure and domain-specific tagging schemas are
provided in Appendix~\ref{app:reference_graph}.

We organize these nodes as a typed directed acyclic graph
\begin{equation}
\mathcal{G}=(\mathcal{V},\mathcal{E}), \qquad
\mathcal{V}=\mathcal{K}\cup\mathcal{R},
\end{equation}
where
\begin{equation}
\mathcal{E}
=
\mathcal{E}_{KK}\cup\mathcal{E}_{KR}\cup\mathcal{E}_{RR}.
\end{equation}
Here, $K\!\rightarrow\!K$ captures scientific dependencies,
$K\!\rightarrow\!R$ links knowledge to reasoning operations that use it, and
$R\!\rightarrow\!R$ captures reasoning dependencies.

To ensure rigorous scientific grounding, we retain only short relation chains
supported within the same reference context, rather than all
graph-theoretically valid paths. Each node, edge, and path remains linked to
its source evidence.

Table~\ref{tab:graph_statistics} summarizes the resulting Biology and
Chemistry graphs. We construct each graph from curated references in its
respective domain, providing domain-specific knowledge and scientific
reasoning evidence.


\subsection{Grounding Reasoning in Scientific Knowledge}
\label{sec:knowledge_reasoning}

Knowledge captures \emph{what scientific content} is present, whereas
reasoning captures \emph{how that knowledge is analyzed or used to support a
scientific conclusion}. This distinction matters because operations such as
comparison, quantitative calculation, causal analysis, and evidence synthesis
are not inherently scientific; their relevance depends on the scientific
knowledge to which they are applied.
\begin{table}[h]
\centering
\small
\begin{tabular}{p{0.94\linewidth}}
\toprule
\textbf{Takeaway 1.}
Reasoning becomes scientifically informative only when grounded in
domain-specific knowledge; $K\!\rightarrow\!R$ explicitly represents this
grounding.\\
\bottomrule
\end{tabular}
\end{table}
SciGraphSelect represents this dependency with directed
$K\!\rightarrow\!R$ relations. For example,
\begin{equation*}
\texttt{stoichiometric constraints}
\;\xrightarrow{\texttt{supports}}\;
\texttt{quantitative reasoning}
\end{equation*}
captures chemistry-grounded reasoning rather than generic arithmetic, while
\begin{equation*}
\texttt{NMR equivalence}
\;\xrightarrow{\texttt{enables}}\;
\texttt{structure elucidation}
\end{equation*}
captures the use of spectroscopic knowledge for molecular inference.

Directionality is preserved in the reference graph and used during document
scoring. During graph construction, a directed relation is retained only when
the scientific reference ssupports both participating nodes and their semantic dependency. Thus, $K_i\!\rightarrow\!R_j$ and
$R_j\!\rightarrow\!K_i$ constitute distinct reference-graph relations.

For a candidate document, however, SciGraphSelect does not require textual
entailment of the directed dependency. It measures relation co-support and
separately evaluates relative directional alignment by contrasting forward and reversed relation verbalizations. Directed edges therefore encode
reference-supported semantic dependencies, while document-side directional
evidence indicates relative consistency with those dependencies rather than
relation-level entailment.
\subsection{Rich Reference Structure, Conservative Document Evidence}
\label{sec:structure_evidence}
\begin{table}[h]
\centering
\small
\begin{tabular}{p{0.94\linewidth}}
\toprule
\textbf{Takeaway 2.}
SciGraphSelect preserves relations supported by scientific references, while
making only conservative claims from candidate-document evidence.\\
\bottomrule
\end{tabular}
\end{table}

The reference graph retains concept dependencies, knowledge--reasoning links,
reasoning dependencies, and their provenance. These relations define which
combinations of scientific concepts and reasoning operations are structurally
meaningful.

Candidate documents provide weaker evidence than the references from which the
graph is constructed. SciGraphSelect therefore uses \emph{relation co-support}
to measure whether both nodes participating in a reference relation are
supported by the document, without assuming that the document entails the
relation itself. Directional alignment separately compares the candidate
representation against forward and reversed relation verbalizations, using
their score difference as a relative consistency signal rather than as
relation-level entailment evidence.

Thus, reference structure specifies \emph{which scientific relations to look
for}, while document scoring measures only the evidence that can be
conservatively supported by the candidate text.

\subsection{Graph-Consistent Co-support and Directional Alignment}
\label{sec:coactivation}

Let $\mathbf{h}_x$ denote the embedding of candidate document $x$, and
$\mathbf{h}_{q_v}$ the embedding of graph node $v$. We define the support of
node $v$ using its activation threshold $\tau_v$:
\begin{equation}
s_v(x)
=
\operatorname{clip}_{[0,1]}
\left(
\frac{
\cos(\mathbf{h}_x,\mathbf{h}_{q_v})-\tau_v
}{
1-\tau_v
}
\right).
\label{eq:node_support}
\end{equation}
Knowledge and reasoning coverage, $S_K(x)$ and $S_R(x)$, aggregate these
supports over knowledge and reasoning nodes, respectively.

For each directed reference edge $(u\!\rightarrow\!v)$, we assign a
reference-derived weight
\begin{equation}
\widetilde{w}(u,v)
=
\underbrace{\bar{c}(u,v)}_{\text{mean confidence}}
\cdot
\underbrace{
\left(
\frac{n(u,v)}{n(v)}
\right)^{\rho}
}_{\text{conditional support}}
\cdot
\underbrace{
\log\!\left(1+n(u,v)\right)
}_{\text{support frequency}},
\label{eq:raw_edge_weight}
\end{equation}

where $n(u,v)$ is the number of reference attestations of the edge,
$n(v)$ is the reference support of its target node,
$\bar{c}(u,v)$ is the mean annotation confidence, and $\rho$ controls
the conditional-support term.
Thus, incoming edges to the same target are normalized to sum to one.
The document-level evidence for a reference edge is
\begin{equation}
e_x(u,v)
=
\min\!\left(s_u(x),s_v(x)\right)w(u,v), \quad w(u,v)
=
\frac{\widetilde{w}(u,v)}
{\sum_{u':(u'\rightarrow v)\in\mathcal{E}}
\widetilde{w}(u',v)}
\label{eq:edge_support}
\end{equation}
Similarly, for a reference-attested path
$p=(v_1,\ldots,v_m)$, we define
\begin{equation}
p_x(p)
=
\min_{i=1,\ldots,m}s_{v_i}(x).
\label{eq:path_support}
\end{equation}
The minimum acts as a conjunctive bottleneck: both nodes participating in a reference relation, or
all constituents of a path, must be supported. These quantities measure
reference-consistent co-support and do not imply that the candidate document
entails the corresponding relation or reasoning sequence.

Because edge co-support itself does not distinguish $u\!\rightarrow\!v$ from
$v\!\rightarrow\!u$, we model directionality separately. For each asymmetric
reference edge, let $q^+_{u\rightarrow v}$ denote its forward verbalization
and $q^-_{u\rightarrow v}$ its reversed counterpart. We define
\begin{equation}
d_x(u\!\rightarrow\!v)
=
\cos\!\left(
\mathbf{h}_x,\mathbf{h}_{q^+_{u\rightarrow v}}
\right)
-
\cos\!\left(
\mathbf{h}_x,\mathbf{h}_{q^-_{u\rightarrow v}}
\right).
\label{eq:directional_margin}
\end{equation}
A positive margin indicates stronger alignment with the reference direction
than with its reversal. The reference edge weight $w(u,v)$ is used only for
edge co-support and is not applied to this directional margin.

SciGraphSelect aggregates node, edge, path, and directional evidence into
document-level scores. The aggregation and final score composition differ
slightly between Biology and Chemistry and are therefore specified in
Appendix~\ref{app:scoring_details}.

\section{Evaluation}
\label{sec:evaluation}

We evaluate SciGraphSelect on Biology and Chemistry in terms of
\emph{domain enrichment}, \emph{structural contribution}, and
\emph{attribution faithfulness}.
We consider both scientific-domain selection within PeS2o~\citep{peS2o} and
broader LLM data selection using a heterogeneous corpus containing scientific,
reasoning, web, and general-domain documents.

We compare \textsc{SciGraphSelect} with semantic similarity, and
progressively structured variants under the same document representation.
Performance is measured with top-budget precision and ranking metrics, while
component analysis examines how knowledge, reasoning, relation, path, and
directional evidence alter the selected corpus.

We ask three questions:
\textbf{RQ1}---Does SciGraphSelect improve target-domain enrichment?
\textbf{RQ2}---How does each structural signal affect selection?
\textbf{RQ3}---Are the resulting knowledge--reasoning attributions supported
by the source documents?
Evaluation details and the human-attribution evaluation are provided in
Appendix~\ref{app:evaluation_details}.

\section{Results}
\label{sec:results}

\subsection{Does Direction-Aware Structure Improve Scientific Data Selection?}
\label{sec:selection_results}

\begin{table*}[t]
\centering
\scriptsize
\setlength{\tabcolsep}{2.8pt}
\renewcommand{\arraystretch}{0.93}
\caption{
Scientific document selection under heterogeneous hard negatives and the
15.1M PeS2o scientific-paper pool.
Chemistry uses the largest directional margin, while Biology averages the
three largest margins.
Bold indicates the best result within each domain, setting, and metric.
}
\label{tab:combined_selection_results}

\resizebox{\textwidth}{!}{%
\begin{tabular}{lccccc|ccccc}
\toprule
&
\multicolumn{5}{c|}{\textbf{Heterogeneous Hard-Negative Validation}}
&
\multicolumn{5}{c}{\textbf{PeS2o Scientific-Paper Evaluation}}
\\
\cmidrule(lr){2-6}
\cmidrule(lr){7-11}

\textbf{Method}
& \textbf{P@1\%}
& \textbf{P@5\%}
& \textbf{P@10\%}
& \textbf{AUPRC}
& \textbf{AUROC}
& \textbf{P@1\%}
& \textbf{P@5\%}
& \textbf{P@10\%}
& \textbf{AUPRC}
& \textbf{AUROC}
\\
\midrule

\multicolumn{11}{l}{\textbf{\textit{Chemistry}}} \\
\addlinespace[1pt]

\quad Embedding similarity
& .7681 & .8197 & .7621 & .727481 & .956629
& .8229 & .7139 & .5833 & .601362 & .921411
\\

\quad Concept only
& .7965 & .8666 & \textbf{.7997}
& .755068 & .957812
& .8921 & .7722 & .6129 & .647915 & .923127
\\

\quad + Typed relation co-activation
& .7992 & .8629 & .7904 & .744855 & .953926
& .9046 & .7796 & .6172 & .655911 & .923901
\\

\quad + Reasoning/path
& .7951 & .8442 & .7657 & .719709 & .948361
& .9000 & .7749 & .6159 & .653231 & .924460
\\

\quad Structure-aware
& .8021 & .8527 & .7824 & .740740 & .954520
& .9000 & .7760 & .6180 & .655088 & .924872
\\

\quad Directional alignment
& .7982 & .5243 & .3948 & .364093 & .752904
& .6838 & .3864 & .2761 & .253820 & .648587
\\

\quad \textbf{Final direction-aware}
& \textbf{.8753}
& \textbf{.8714}
& .7829
& \textbf{.767914}
& \textbf{.959490}
& \textbf{.9635}
& \textbf{.7992}
& \textbf{.6191}
& \textbf{.681961}
& \textbf{.926256}
\\

\midrule

\multicolumn{11}{l}{\textbf{Biology}} \\
\addlinespace[1pt]

\quad Embedding similarity
& .2519 & .5358 & .6003 & .533929 & .936527
& .8780 & .8400 & .8123 & .739965 & .874480
\\

\quad Concept only
& .2374 & .5644 & .6266 & .552135 & .940113
& .9111 & .8752 & .8466 & .767407 & .884933
\\

\quad + Typed relation co-activation
& .2381 & .5647 & .6270 & .552377 & .940144
& .9115 & .8754 & .8467 & .767476 & .884942
\\

\quad + Reasoning/path
& .2387 & .5654 & .6266 & .553505 & .941269
& .9116 & .8752 & .8464 & .768633 & .886166
\\

\quad Structure-aware
& .2387 & .5654 & .6268 & .552621 & .940170
& .9116 & .8752 & .8464 & .767463 & .885009
\\

\quad Directional alignment
& .3408 & .3128 & .3030 & .266727 & .767182
& .8368 & .7689 & .7319 & .643158 & .799965
\\

\quad \textbf{Final direction-aware}
& \textbf{.6701}
& \textbf{.7313}
& \textbf{.6966}
& \textbf{.664223}
& \textbf{.954632}
& \textbf{.9405}
& \textbf{.9129}
& \textbf{.8865}
& \textbf{.811951}
& \textbf{.908901}
\\

\bottomrule
\end{tabular}%
}
\vspace{-2mm}
\end{table*}

Table~\ref{tab:combined_selection_results} reveals a clear complementarity
between structural relevance and directional evidence.
Concept-based scoring provides a strong relevance signal, while relation co-support and reasoning/path components offer smaller, domain-dependent refinements.
Directional alignment alone, however, performs poorly as a general ranking
criterion, indicating that directional consistency is not sufficient to
identify scientifically relevant documents.

The picture changes when directional alignment is combined with the
structure-aware score.
Final Direction-Aware consistently yields the strongest selection across both
Biology and Chemistry and under both scientific-paper and heterogeneous
negative settings, with only a minor exception at the broader Chemistry
hard-negative budget.
The particularly large gain under heterogeneous negatives suggests that
directional evidence helps distinguish target-domain scientific structure from
superficially similar or reasoning-heavy distractors, but only when anchored by
reference-grounded relevance.

Overall, these results support \emph{Takeaway 1}: directional evidence is
useful for scientific data selection when grounded in structured relevance,
rather than when used as an independent selection criterion.










\subsection{What Does Each Structural Component Add?}
\label{sec:stage-transitions}

Aggregate metrics do not reveal how each scoring stage changes the selected
corpus. We therefore compare consecutive top-1\% sets, where \emph{entered}
and \emph{exited} denote documents added to or removed from the selection.
TP/FP follow PeS2o source-domain labels, so positive $\Delta$TP indicates more
target-domain documents at a fixed budget. Turnover measures set change rather
than quality.

\begin{table}[t]
\centering
\small
\caption{
Top-1\% selection transitions over the common PeS2o pool.
TP/FP are defined using PeS2o source-domain proxy labels.
Arrows indicate the preferred direction when applicable; turnover measures
set change rather than quality.
}
\label{tab:top1-transition}
\resizebox{\linewidth}{!}{%
\begin{tabular}{lrrrrr}
\toprule
\textbf{Transition}
& \textbf{Retained}
& \textbf{Entered TP$\uparrow$/FP$\downarrow$}
& \textbf{Exited TP$\downarrow$/FP$\uparrow$}
& \textbf{$\Delta$TP$\uparrow$}
& \textbf{Turnover} \\
\midrule

\multicolumn{6}{l}{\textbf{Chemistry}} \\
\quad Embedding Similarity $\rightarrow$ Concept
& 93,352 & 51,259 / 6,513 & 40,800 / 16,972 & +10,459 & 38.23\% \\
\quad Concept $\rightarrow$ Relation co-activation
& 124,607 & 23,670 / 2,847 & 21,783 / 4,734 & +1,887 & 17.55\% \\
\quad Relation co-activation $\rightarrow$ Reasoning/path
& 129,516 & 18,202 / 3,406 & 18,897 / 2,711 & $-$695 & 14.30\% \\
\quad Structure-aware $\rightarrow$ Final direction-aware
& 102,992 & 46,732 / 1,400 & 37,137 / 10,995 & +9,595 & 31.85\% \\

\midrule

\multicolumn{6}{l}{\textbf{Biology}} \\
\quad Embedding Similarity $\rightarrow$ Concept
& 108,882 & 38,886 / 3,356 & 33,889 / 8,353 & +4,997 & 27.95\% \\
\quad Concept $\rightarrow$ Relation co-activation
& 149,618 & 1,396 / 110 & 1,338 / 168 & +58 & 1.00\% \\
\quad Relation co-activation $\rightarrow$ Reasoning/path
& 146,135 & 4,527 / 462 & 4,507 / 482 & +20 & 3.30\% \\
\quad Structure-aware $\rightarrow$ Final direction-aware
& 38,818 & 105,080 / 7,226 & 100,711 / 11,595 & +4,369 & 74.31\% \\

\bottomrule
\end{tabular}%
}
\end{table}

Table~\ref{tab:top1-transition} shows that the structural components alter
the selected corpus in different ways. Concept aggregation produces the
largest initial improvement: compared with embedding similarity, it replaces
38.23\% of the Chemistry selection and 27.95\% of the Biology selection,
while increasing the number of target-domain documents by 10,459 and 4,997,
respectively. This indicates that aggregating support from multiple scientific
concepts recovers target-domain documents that are not captured by a single
nearest semantic match.

Relation co-activation provides a smaller and strongly domain-dependent
refinement. It adds 1,887 net Chemistry documents but only 58 Biology
documents, with turnovers of 17.55\% and 1.00\%, respectively. Reasoning/path
evidence changes the selected sets further but does not consistently improve
source-domain precision: it yields only +20 net Biology documents and
$-$695 in Chemistry. Thus, reasoning evidence primarily changes \emph{what
kind of scientific content is selected}, rather than uniformly increasing
target-domain purity.

\begin{figure}[t]
\centering
\includegraphics[width=0.9\linewidth]{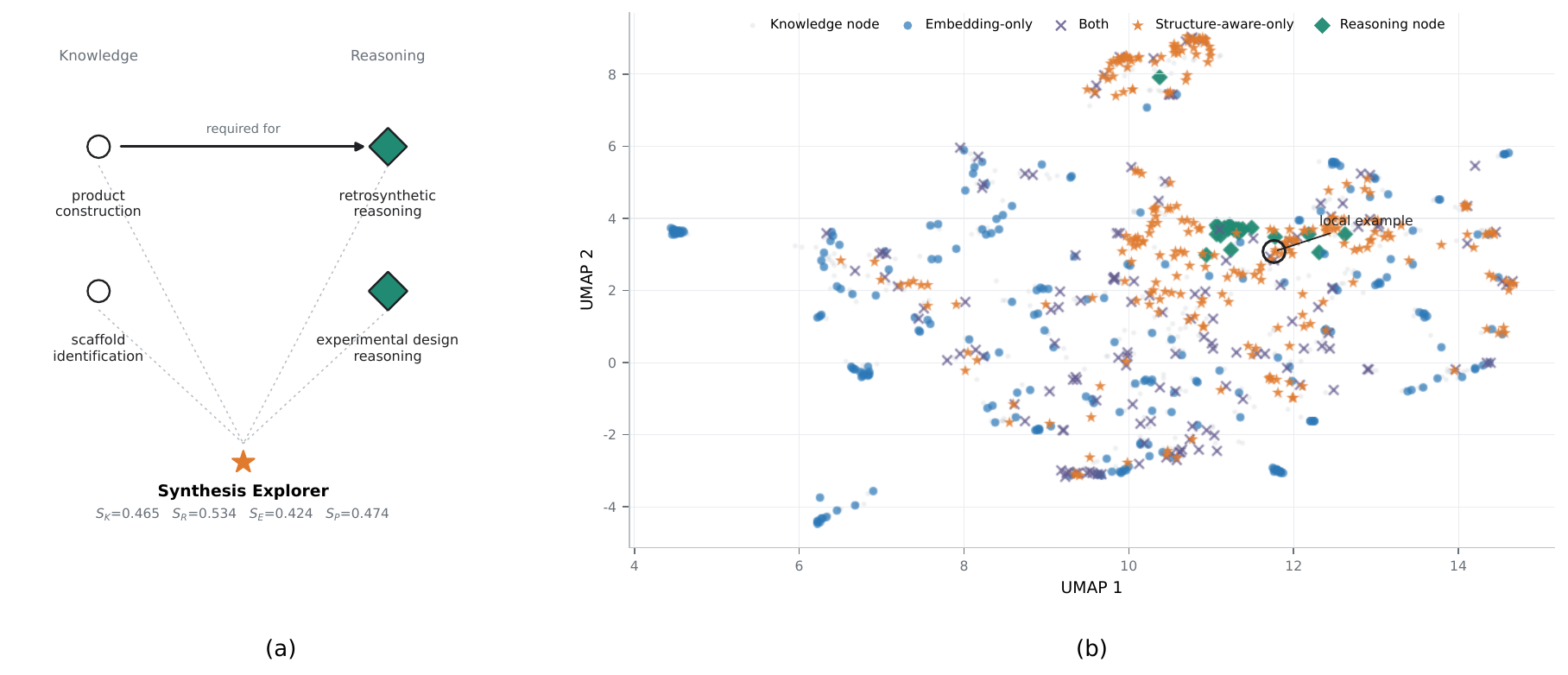}
\caption{
Structured scoring changes both where and what is selected.
(a) Local graph evidence for a Chemistry document selected only by the
structure-aware score.
(b) Joint UMAP of embedding-only, shared, and structure-aware-only selections
with active knowledge and reasoning nodes.
UMAP is used only for visualization; selection and relation similarities are
computed in the original embedding space.
}
\label{fig:structured-selection}
\end{figure}

Figure~\ref{fig:structured-selection} provides a complementary view in
Chemistry. Flat-only and structure-only documents occupy different regions of
the joint document--graph embedding, while the local example illustrates how
a structure-only document receives support from multiple knowledge and
reasoning nodes connected by reference-attested relations. Importantly, this
is graph-consistent co-support rather than evidence that the candidate
document entails the reference relation.

The final transition shows a different effect. Adding directional alignment
to the structure-aware score produces positive $\Delta$TP in both domains:
+9,595 for Chemistry and +4,369 for Biology. The large turnover in Biology
(74.31\%) also shows that directional alignment substantially reorders the
top-ranked set rather than merely adjusting scores within the same selection.
Together with Table~\ref{tab:combined_selection_results}, these results support
\emph{Takeaway 1}: directional evidence is useful when combined with
reference-grounded structural relevance, whereas the intermediate structural
components have distinct and domain-dependent effects.

\subsection{Are the Selections and Their Rationales Human-Verifiable?}
\label{sec:results_human}

The previous analyses rely on PeS2o source-domain labels as proxy relevance
judgments. We therefore complement them with human evaluation of
(i) whether documents uniquely selected by SciGraphSelect are more relevant
than those uniquely selected by embedding similarity and
(ii) whether the reported concept, reasoning, and relation attributions are
supported by the selected documents.

\begin{table}[t]
\centering
\small
\caption{
Human relevance of documents uniquely selected by embedding similarity or
SciGraphSelect. All metrics are higher-is-better.
}
\label{tab:human_relevance}
\begin{tabular}{lccc}
\toprule
\textbf{Domain}
& \textbf{Embedding Similarity-only $\uparrow$}
& \textbf{SciGraphSelect-only $\uparrow$}
& \textbf{$\Delta\uparrow$} \\
\midrule
Chemistry & .821 & .853 & +.032 \\
Biology   & .915 & .963 & +.049 \\
\bottomrule
\end{tabular}
\end{table}

Human judgments favor SciGraphSelect-only documents in both domains,
indicating that the changes in selection are not explained solely by PeS2o
source-domain labels but also correspond to higher semantic relevance.

\begin{table}[t]
\centering
\small
\caption{
Human evaluation of SciGraphSelect attribution correctness.
All metrics are higher-is-better.
}
\label{tab:human_eval}
\begin{tabular}{lccccc}
\toprule
\textbf{Domain}
& \textbf{Rel.$\uparrow$}
& \textbf{Concept$\uparrow$}
& \textbf{Reason.$\uparrow$}
& \textbf{Relation$\uparrow$}
& $\boldsymbol{\kappa}\uparrow$ \\
\midrule
Chemistry & .837 & .524 & .521 & .426 & .691 \\
Biology   & .939 & .890 & .789 & .503 & .719 \\
\bottomrule
\end{tabular}
\end{table}

Table~\ref{tab:human_eval} evaluates whether the reported knowledge,
reasoning, and relation evidence are supported by the document, together with
inter-annotator agreement.
Attributions are generally more reliable in Biology, while relation-level
support is the weakest component in both domains.
This is consistent with our conservative interpretation: cco-support of reference-linked nodes does not guarantee that the corresponding relation
is explicitly expressed in the candidate document.

Overall, the human evaluation shows that SciGraphSelect improves
human-judged relevance while exposing selection rationales whose reliability
can be directly audited.

\section{Discussion and Limitations}
\label{sec:discussion}

Our results support a structured view of scientific data selection.
Concept evidence provides the strongest relevance signal, while relation and
reasoning components reshape the selected corpus, and directional alignment is
most useful when combined with reference-grounded structure.
This suggests that scientific data selection benefits from modeling not only
\emph{what} knowledge is present, but also how it participates in
reference-attested structure.

We design SciGraphSelect to separate rich reference structure from
conservative document evidence.
Relations and paths are reference-supported, whereas candidate documents are
scored from whole-document embeddings; relation co-support therefore does not establish relation entailment or recover a reasoning sequence.
Accordingly, SciGraphSelect measures relevance and structural coverage rather
than factual correctness or downstream utility, and its large-scale evaluation
uses PeS2o source labels as proxy relevance judgments.

Because reference structure and document evidence are separated, the current
scorer can be replaced by finer-grained sentence-, span-, or relation-level
models while reusing the same graph, and the framework can extend to new
domains through domain-matched references.
SciGraphSelect is therefore best viewed as an auditable upstream
data-selection framework that complements downstream scientific verification.
\section{Conclusion}
\label{sec:conclusion}

We introduce \textsc{SciGraphSelect}, a reference-grounded framework for auditable
scientific data selection.
SciGraphSelect represents domain-specific knowledge, reusable reasoning
operations, and reference-attested relations in a typed graph, and uses this
structure to rank candidate documents while restricting document-level claims
to conservative co-support and directional-alignment evidence.

Across Biology and Chemistry, SciGraphSelect improves target-domain
enrichment over embedding similarity and reveals that different structural
signals change the selected corpus in distinct, domain-dependent ways.
Human evaluation further shows that the resulting selections are more
semantically relevant and that their supporting concepts and reasoning
operations can be directly inspected, while also exposing relation-level
attribution as the primary remaining weakness.

By making both the selection score and its supporting scientific evidence
traceable to reference structure, SciGraphSelect extends interpretability from
model outputs to scientific data curation itself.
It provides a modular basis for more transparent construction of scientific
training corpora and for future integration with finer-grained evidence
grounding and downstream scientific verification.



\bibliographystyle{plainnat}
\bibliography{neurips_2026}







\appendix

\section{Reference Graph Construction}
\label{app:reference_graph}

\subsection{Reference Corpus Construction}
\label{app:reference}

SciGraphSelect constructs a separate reference graph for each target domain
from scientific QA and reasoning resources. Each reference provides a
question together with an answer, explanation, or reasoning trace, from which
we extract reusable knowledge concepts, reasoning operations, and supported
relations. The original reference text and source provenance are retained for
all graph elements.

\subsubsection{Chemistry References}
\label{app:chem_reference}

The Chemistry reference corpus is assembled from chemistry-focused benchmark
and scientific-reasoning resources, including HLE Chemistry
\cite{phan2025hle}, ChemQuests \cite{amiri2025chemquests}, WildSci Chemistry
\cite{liu2026wildsci}, and Nemotron Science Chemistry
\cite{nvidia2026nemotronsciencev2}.
References are normalized into question, reasoning, and answer fields before
tagging, and distinct reasoning traces are retained when they provide
different reference-supported structures.
The final Chemistry graph is constructed from 146,871 validated references.

\subsubsection{Biology References}
\label{app:bio_reference}

The Biology reference corpus is constructed from biology-focused scientific
QA and reasoning resources, including GPQA Biology \cite{rein2024gpqa},
HLE Biology \cite{phan2025hle}, AA-Omniscience Biology
\cite{jackson2025aaomniscience}, and Biology examples from Nemotron Science
\cite{nvidia2026nemotronsciencev2}.
References are selected according to their semantic Biology assignment,
yielding 38,217 validated Biology references.

Table~\ref{tab:graph_statistics} summarizes the resulting domain-matched
reference graphs.

\begin{table}[t]
    \centering
    \small
    \caption{Statistics of the domain-matched reference graphs.}
    \label{tab:graph_statistics}
    \begin{tabular}{lrr}
        \toprule
        & \textbf{Biology} & \textbf{Chemistry} \\
        \midrule
        References       & 38,217  & 146,871 \\
        Knowledge nodes  & 35,822  & 197,462 \\
        Reasoning nodes  & 20      & 23 \\
        Backbone edges   & 5,716   & 53,686 \\
        Attested paths   & 6,320   & 56,711 \\
        \bottomrule
    \end{tabular}
\end{table}
\subsection{Reference Tagging}
\label{app:tagging}

\subsubsection{Tagging Model and Input}
\label{app:tag_model}

Reference tagging is performed with GPT-4o-mini
\cite{openai2024gpt4omini}. The tagger receives the scientific question together with the available
answer, explanation, or reasoning trace. Biology and Chemistry use separate
graph prompts tailored to their domain taxonomies and controlled
reasoning vocabularies; both prompts require reusable knowledge and reasoning
items rather than question-specific surface forms.

\subsubsection{Structured Tagging Outputs}
\label{app:tag_schema}

Both domains return hierarchical domain information, reusable knowledge
concepts, controlled reasoning operations, supported links among extracted
elements, difficulty, and confidence. Biology additionally records tagging
status, domain relevance/specificity, canonical definitions and aliases, and a
reasoning-basis field. Chemistry separately records knowledge--reasoning links
and reasoning-to-reasoning edges. In both cases, edges are retained only when
the supplied reference supports the corresponding dependency, and unsupported
links are represented by empty arrays rather than inferred from co-occurrence.
\subsubsection{Graph Tagging Prompts}
\label{app:tag_prompts}

The graph prompts are summarized below, preserving the task definition,
domain-specific taxonomy, controlled reasoning vocabulary, relation constraints,
and output schema used for reference tagging.

\begin{promptbox}{Prompt for Chemistry reference tagging}
\scriptsize

\textbf{Task.}
Analyze the \texttt{QUESTION}, \texttt{THINKING}, and \texttt{ANSWER}
together and extract the minimum reusable skills needed to produce the answer.
Do not solve the item again. Avoid copying question-specific compounds,
constants, spectra peak lists, identifiers, or long reaction strings unless
they are stable textbook concepts.

\tcbline
\textbf{Skill types.}
\begin{itemize}[leftmargin=1.5em, itemsep=0.15em, topsep=0.15em]
    \item \texttt{knowledge\_skills}: reusable chemical facts, concepts,
    mechanisms, structures, properties, procedures, and analytical methods.
    \item \texttt{reasoning\_skills}: reusable operations applied to that
    knowledge.
\end{itemize}

\tcbline
\textbf{Domain path.}
Return one primary Chemistry path, typically 2--4 levels, using stable
textbook taxonomy (e.g., Organic Chemistry / Reaction Mechanisms,
Physical Chemistry / Thermodynamics, Analytical Chemistry / NMR Spectroscopy).

\tcbline
\textbf{Knowledge skills.}
Return 1--6 concise reusable noun phrases at a stable textbook granularity.
Allowed types include
\texttt{fact}, \texttt{concept}, \texttt{mechanism}, \texttt{law},
\texttt{rule}, \texttt{structure}, \texttt{property}, \texttt{procedure},
\texttt{analytical\_method}, and \texttt{experimental\_principle}.

\tcbline
\textbf{Reasoning skills.}
Return 1--5 items from the controlled vocabulary, including
\texttt{reaction\_mechanism\_reasoning},
\texttt{stoichiometric\_reasoning},
\texttt{acid\_base\_reasoning},
\texttt{redox\_electron\_accounting},
\texttt{thermodynamic\_reasoning},
\texttt{quantitative\_calculation},
\texttt{spectroscopic\_structure\_elucidation},
\texttt{data\_interpretation}, and related operations.

\tcbline
\textbf{Relations.}
\texttt{skill\_links} connect listed knowledge and reasoning skills using only
directly supported relations such as
\texttt{prerequisite\_of}, \texttt{part\_of}, \texttt{enables},
\texttt{applies\_to}, and \texttt{combines\_with}.
\texttt{reasoning\_edges} capture supported dependencies among reasoning
operations; co-occurrence alone does not justify an edge.

\tcbline
\textbf{Output.}
Return valid JSON containing
\texttt{domain\_path}, \texttt{knowledge\_skills},
\texttt{reasoning\_skills}, \texttt{skill\_links},
\texttt{reasoning\_edges}, \texttt{difficulty}, and \texttt{confidence}.
Unsupported links are represented by empty arrays.
\end{promptbox}

\begin{promptbox}{Prompt for Biology reference tagging}
\scriptsize

\textbf{Task.}
Analyze the question, any explicit thinking or explanation, and the answer
together. Extract the minimum reusable biological knowledge concepts and
reasoning operations needed to produce the answer without solving the item
again.

\tcbline
\textbf{Tagging goal.}
Use stable canonical concept names, separate scientific concepts from reasoning
operations, disambiguate aliases and homonyms, and retain only relations
supported by the supplied reference.

\tcbline
\textbf{Domain path.}
Return one Biology path under Life Sciences using a stable subdomain taxonomy,
such as Cell Biology, Molecular Biology, Genetics, Biochemistry, Microbiology,
Immunology, Neuroscience, Ecology, Physiology, or Anatomy.

\tcbline
\textbf{Knowledge skills.}
Return 1--6 reusable canonical concepts, entities, mechanisms, structures,
methods, or principles. Each item contains a canonical \texttt{name},
\texttt{type}, short \texttt{definition}, optional \texttt{aliases}, and
\texttt{domain\_specificity}. Avoid question-specific values and generic facet
phrases such as ``features of'' or ``role of.''

\tcbline
\textbf{Knowledge relations.}
Return only directly supported relations among listed knowledge skills, using
relations such as
\texttt{is\_a}, \texttt{part\_of}, \texttt{prerequisite\_of},
\texttt{causes}, \texttt{regulates}, \texttt{activates},
\texttt{inhibits}, \texttt{associated\_with},
\texttt{genetic\_basis\_of}, and \texttt{measures}.

\tcbline
\textbf{Reasoning skills.}
Return 1--5 operations from the controlled vocabulary, including
\texttt{mechanistic\_reasoning},
\texttt{genotype\_phenotype\_inference},
\texttt{sequence\_structure\_function\_reasoning},
\texttt{experimental\_result\_interpretation},
\texttt{comparative\_reasoning},
\texttt{multi\_hop\_integration}, and related operations.

\tcbline
\textbf{Links and reasoning edges.}
\texttt{knowledge\_reasoning\_links} connect knowledge concepts to reasoning
operations using \texttt{required\_for}.
\texttt{reasoning\_edges} encode supported dependencies using
\texttt{prerequisite\_of}, \texttt{precedes}, \texttt{enables},
\texttt{verifies}, or \texttt{combines\_with}.

\tcbline
\textbf{Output.}
Return valid JSON containing domain information, knowledge skills and
relations, reasoning skills, knowledge--reasoning links, reasoning edges,
reasoning basis, difficulty, and confidence. Unsupported elements are
represented by empty arrays.
\end{promptbox}
\subsection{Knowledge Canonicalization}
\label{app:canonicalization}

Raw LLM-extracted knowledge phrases can contain lexical variants, punctuation
differences, aliases, or near-duplicate descriptions. Knowledge strings are
case-folded and normalized for dash, whitespace, and punctuation variation,
and candidate near-duplicates are retrieved using Qwen3-Embedding-0.6B \cite{zhang2025qwen3embedding}.
Automatic merging is deliberately conservative: a candidate pair must satisfy
all of the following conditions:
\begin{itemize}
    \item cosine similarity $\geq 0.99$;
    \item mutual top-$k$ neighborhood;
    \item compatible knowledge types;
    \item token Jaccard similarity $\geq 0.80$;
    \item sequence similarity ratio $\geq 0.85$;
    \item matching numerical tokens when numbers are present; and
    \item complete-link cluster cosine $\geq 0.99$.
\end{itemize}
Pairs with cosine similarity above $0.92$ that fail the automatic merge
criteria are retained as review/family candidates rather than aggressively
collapsed. For Chemistry, 198,214 extracted knowledge nodes are reduced to
197,462 canonical nodes, with 764 automatically merged pairs.

\subsection{Typed DAG Construction and Reference Weights}
\label{app:graph_construction}

After canonicalization, we construct
\begin{equation}
\mathcal{G}=(\mathcal{V},\mathcal{E}),
\qquad
\mathcal{V}=\mathcal{V}_K\cup\mathcal{V}_R,
\qquad
\mathcal{E}=\mathcal{E}_{KK}\cup\mathcal{E}_{KR}\cup\mathcal{E}_{RR}.
\end{equation}
Here, $\mathcal{E}_{KK}$ represents dependencies between scientific knowledge
concepts, $\mathcal{E}_{KR}$ links domain knowledge to reasoning operations,
and $\mathcal{E}_{RR}$ represents dependencies between reusable reasoning
operations. Cycles are removed from the directed backbone. We retain only
short paths explicitly attested within an individual reference rather than all
paths implied by graph connectivity.

For a directed edge $(u\!\rightarrow\!v)$, reference evidence determines
\begin{equation}
\widetilde{w}(u,v)
=
\underbrace{\bar{c}(u,v)}_{\text{mean confidence}}
\cdot
\underbrace{\left(\frac{n(u,v)}{n(v)}\right)^{\rho}}_{\text{conditional support}}
\cdot
\underbrace{\log\!\left(1+n(u,v)\right)}_{\text{support frequency}},
\end{equation}
followed by incoming normalization
\begin{equation}
w(u,v)
=
\frac{\widetilde{w}(u,v)}
{\sum_{u':(u'\rightarrow v)\in\mathcal{E}}\widetilde{w}(u',v)}.
\end{equation}

\begin{table}[t]
    \centering
    \small
    \caption{Composition of the Chemistry typed graph.}
    \label{tab:chem_edge_stats}
    \begin{tabular}{lr}
        \toprule
        \textbf{Graph element} & \textbf{Count} \\
        \midrule
        Knowledge nodes & 197,462 \\
        Reasoning nodes & 23 \\
        Backbone edges & 53,686 \\
        $K\!\rightarrow\!K$ edges & 12 \\
        $K\!\rightarrow\!R$ edges & 53,650 \\
        $R\!\rightarrow\!R$ edges & 24 \\
        Attested paths & 56,711 \\
        \midrule
        $K\!-\!R$ paths & 53,650 \\
        $R\!-\!R$ paths & 24 \\
        $K\!-\!R\!-\!R$ paths & 3,028 \\
        $K\!-\!K\!-\!R$ paths & 9 \\
        \bottomrule
    \end{tabular}
\end{table}

The graph represents reference-attested scientific structure rather than a
recovered document-level reasoning trace.

\section{Scoring Details}
\label{app:scoring_details}

Section~\ref{sec:coactivation} defines the common node activation, edge
co-support, path bottleneck, and directional-margin signals.  Here we specify
how these signals are aggregated in each domain.  We use
$S_{\mathrm{graph}}$ to denote the domain-specific structure-aware score
obtained before ECDF normalization and directional fusion, and
$S_{\mathrm{final}}$ to denote the final direction-aware score.

\subsection{Common Graph-Evidence Components}

Let $s_v(x)$ denote the thresholded node activation defined in the main text.
We partition edges into the knowledge-anchored base group
$\mathcal{E}_{\mathrm{base}}=\mathcal{E}_{KK}\cup\mathcal{E}_{KR}$ and the
reasoning-only group $\mathcal{E}_{RR}$.  Paths are analogously separated into
base paths and paths containing an $R\!\rightarrow\!R$ extension.  This
separation allows reasoning-only structure to supplement the score while
retaining knowledge-anchored evidence as its base.

\subsection{Chemistry Scoring}
\label{app:chem_scoring}

\paragraph{Knowledge and reasoning coverage.}
Let $\mathcal{K}_{16}(x)$ and $\mathcal{R}_{5}(x)$ denote the highest activated
knowledge and reasoning nodes, respectively, up to the indicated budget.
\begin{align}
S_K^{\mathrm{Chem}}(x)
&=
\frac{1}{|\mathcal{K}_{16}(x)|}
\sum_{v\in\mathcal{K}_{16}(x)}a_v(x),\\
S_R^{\mathrm{Chem}}(x)
&=
\frac{1}{|\mathcal{R}_{5}(x)|}
\sum_{v\in\mathcal{R}_{5}(x)}a_v(x).
\end{align}
The score is zero when the corresponding active set is empty.

\paragraph{Edge and path co-support.}
For each $g\in\{\mathrm{base},RR\}$, Chemistry computes a top-16
weight-normalized edge mean,
\begin{equation}
S_{E,g}^{\mathrm{Chem}}(x)
=
\frac{
\sum_{e\in\operatorname{Top16}_x(\mathcal{E}_g)}e_x(e)
}{
\sum_{e\in\operatorname{Top16}_x(\mathcal{E}_g)}w(e)
},
\end{equation}
and an unweighted top-16 path mean,
\begin{equation}
S_{P,g}^{\mathrm{Chem}}(x)
=
\frac{1}{|\operatorname{Top16}_x(\mathcal{P}_g)|}
\sum_{p\in\operatorname{Top16}_x(\mathcal{P}_g)}p_x(p).
\end{equation}
The ungated component summaries are
$S_E^{\mathrm{Chem}}=\max(S_{E,\mathrm{base}},S_{E,RR})$ and
$S_P^{\mathrm{Chem}}=\max(S_{P,\mathrm{base}},S_{P,RR})$.

\paragraph{Domain gating, penalties, and graph score.}
Chemistry gates generic reasoning and $R\!\rightarrow\!R$ structure using
\begin{equation}
g_{\mathrm{Chem}}(x)
=
\min\!\left(1,\frac{A_{\mathrm{Chem}}(x)}{0.20}\right).
\end{equation}
The gated components are
\begin{align}
\widetilde{S}_R(x)
&=
g_{\mathrm{Chem}}(x)S_R(x),\\
\widetilde{S}_E(x)
&=
\max\!\left(
S_{E,\mathrm{base}}(x),
g_{\mathrm{Chem}}(x)S_{E,RR}(x)
\right),\\
\widetilde{S}_P(x)
&=
\max\!\left(
S_{P,\mathrm{base}}(x),
g_{\mathrm{Chem}}(x)S_{P,RR}(x)
\right).
\end{align}
The pre-penalty aggregate is
\begin{equation}
S_{\mathrm{pre}}^{\mathrm{Chem}}(x)
=
0.50S_K(x)
+0.15\widetilde{S}_R(x)
+0.20\widetilde{S}_E(x)
+0.15\widetilde{S}_P(x).
\end{equation}
The cross-domain penalty is
\begin{equation}
C(x)
=
0.15A_{\mathrm{cross}}(x)
\left(1-g_{\mathrm{Chem}}(x)\right).
\end{equation}
Let $T(x)$ denote document token count and let $M_{\mathrm{frag}}(x)$ indicate
whether the document matches a query- or title-fragment pattern.  The fragment
penalty is
\begin{equation}
F_{\mathrm{frag}}(x)
=
0.10\,
\mathbb{I}
\left[
T(x)<96
\;\lor\;
M_{\mathrm{frag}}(x)=1
\right].
\end{equation}
The fragment patterns detect search-filter expressions, search queries or
strategies, keyword lists, and strings beginning with title-, objective-, aim-,
or goal-style metadata markers.  The Chemistry graph score is
\begin{equation}
S_{\mathrm{graph}}^{\mathrm{Chem}}(x)
=
\operatorname{clip}_{[0,1]}
\left(
S_{\mathrm{pre}}^{\mathrm{Chem}}(x)
-C(x)
-F_{\mathrm{frag}}(x)
\right).
\label{eq:chem_graph_score}
\end{equation}

\subsection{Biology Scoring}
\label{app:bio_scoring}

Biology uses monotone pooling.  For values sorted as
$z_{(1)}\geq\cdots\geq z_{(L)}$, with zero-padding when fewer than $L$ values
are available,
\begin{equation}
\operatorname{MPool}_L(z)
=
\frac12 z_{(1)}
+
\frac12\frac{1}{L}\sum_{i=1}^{L}z_{(i)}.
\end{equation}

\paragraph{Knowledge and reasoning coverage.}
Knowledge nodes are grouped using the complete-link family mapping, and each
family contributes its maximum activation:
\begin{equation}
S_K^{\mathrm{Bio}}(x)
=
\operatorname{MPool}_8
\left(
\left\{
\max_{v\in F}a_v(x)
:
F\in\mathcal{F}_K(x)
\right\}
\right).
\end{equation}
Reasoning is pooled and then knowledge-gated:
\begin{align}
S_R^{\mathrm{raw}}(x)
&=
\operatorname{MPool}_3
\left(
\{a_r(x):r\in\mathcal{R}\}
\right),\\
g_K(x)
&=
\min\!\left(1,\frac{S_K(x)}{0.15}\right),\\
S_R^{\mathrm{Bio}}(x)
&=
g_K(x)S_R^{\mathrm{raw}}(x).
\end{align}

\paragraph{Edge co-support.}
For each target $t$, incoming weighted evidence is accumulated as
\begin{equation}
z_t^E(x)
=
\min\!\left(
1,
\sum_{u:(u\rightarrow t)\in\mathcal{E}}
\min(a_u(x),a_t(x))w(u,t)
\right).
\end{equation}
Base and $R\!\rightarrow\!R$ evidence are pooled separately:
\begin{align}
S_{E,\mathrm{base}}(x)
&=
\operatorname{MPool}_8
\left(
\{z_t^E(x)\}_{\mathrm{base}}
\right),\\
S_{E,RR}^{\mathrm{raw}}(x)
&=
\operatorname{MPool}_8
\left(
\{z_t^E(x)\}_{RR}
\right).
\end{align}
The reasoning-extended edge component is
\begin{equation}
S_E(x)
=
S_{E,\mathrm{base}}(x)
+
0.20
\min\!\left(
S_{E,\mathrm{base}}(x),
S_{E,RR}^{\mathrm{raw}}(x)
\right)
\left(
1-S_{E,\mathrm{base}}(x)
\right).
\end{equation}

\paragraph{Path co-support.}
Biology additionally weights path evidence by reference support reliability,
\begin{equation}
r(n_p)=1-\exp(-n_p/2),
\end{equation}
and forms a terminal-wise score
\begin{equation}
z_t^P(x)
=
\max_{p:\operatorname{terminal}(p)=t}
\left[
\min_{v\in p}a_v(x)\,r(n_p)
\right].
\end{equation}
Base and $RR$-extended path evidence are separately pooled with
$\operatorname{MPool}_8$, giving $S_{P,\mathrm{base}}$ and
$S_{P,RR}^{\mathrm{raw}}$.  Their combined path component is
\begin{equation}
S_P(x)
=
S_{P,\mathrm{base}}(x)
+
0.20
\min\!\left(
S_{P,\mathrm{base}}(x),
S_{P,RR}^{\mathrm{raw}}(x)
\right)
\left(
1-S_{P,\mathrm{base}}(x)
\right).
\end{equation}

\paragraph{Biology graph score.}
The knowledge-anchored typed-edge core is
\begin{equation}
S_{\mathrm{PT}}^{\mathrm{Bio}}(x)
=
S_K(x)
+
\left(1-S_K(x)\right)
0.10S_{E,\mathrm{base}}(x),
\end{equation}
and the full graph score is
\begin{equation}
\begin{split}
S_{\mathrm{graph}}^{\mathrm{Bio}}(x)
=
S_{\mathrm{PT}}^{\mathrm{Bio}}(x)
+
\left(
1-S_{\mathrm{PT}}^{\mathrm{Bio}}(x)
\right)
\left(
0.08S_R(x)
+0.07S_E(x)
+0.05S_P(x)
\right).
\end{split}
\label{eq:bio_graph_score}
\end{equation}
Thus, the $0.10$ edge term forms the knowledge-anchored typed-edge core,
whereas the later $0.07$ term incorporates the reasoning-extended edge
component.

\subsection{Directional Alignment and Fusion}
\label{app:direction_fusion}

Let $d_{(1)}(x)\geq d_{(2)}(x)\geq\cdots$ denote the sorted
forward-minus-reverse margins over asymmetric reference edges.  We aggregate
\begin{equation}
D_k(x)
=
\frac{1}{k}
\sum_{i=1}^{k}d_{(i)}(x).
\end{equation}
For each domain $d\in\{\mathrm{Bio},\mathrm{Chem}\}$, let
$\mathcal{C}_d$ denote the fixed domain-specific validation pool.
We separately percentile-normalize the graph score and directional
margin using their empirical CDFs on $\mathcal{C}_d$, denoted by
$\Pi_{d,\mathrm{graph}}$ and $\Pi_{d,\mathrm{dir}}$, respectively.

Biology uses $k=3$, whereas Chemistry uses $k=1$.  Reference edge weights
$w(u,v)$ are not applied to these margins.  Directional fusion is applied only
after the domain-specific graph score has been fully computed; Chemistry's
domain gate and penalties are therefore incorporated before fusion:
\begin{align}
S_{\mathrm{final}}^{\mathrm{Bio}}(x)
&=
0.7\Pi_{\mathrm{Bio},\mathrm{graph}}
\left(
S_{\mathrm{graph}}^{\mathrm{Bio}}(x)
\right)
+
0.3\Pi_{\mathrm{Bio},\mathrm{dir}}
\left(
D_3^{\mathrm{Bio}}(x)
\right),\\
S_{\mathrm{final}}^{\mathrm{Chem}}(x)
&=
0.9\Pi_{\mathrm{Chem},\mathrm{graph}}
\left(
S_{\mathrm{graph}}^{\mathrm{Chem}}(x)
\right)
+
0.1\Pi_{\mathrm{Chem},\mathrm{dir}}
\left(
D_1^{\mathrm{Chem}}(x)
\right).
\end{align}

\subsection{Hyperparameter Selection}
\label{app:hyperparameters}

The directional-fusion weight is selected on the heterogeneous hard-negative
validation set using AUPRC from
$\eta\in\{0,0.1,0.2,0.3,0.5\}$.  The reported settings use
$\eta_{\mathrm{Bio}}=0.3$ with top-3 directional aggregation and
$\eta_{\mathrm{Chem}}=0.1$ with top-1 aggregation.  These settings are fixed
before evaluation on the full PeS2o scientific-paper pool.

\begin{table}[t]
\centering
\small
\caption{Domain-specific graph-score aggregation and directional-fusion
settings used by SciGraphSelect.}
\label{tab:domain_scoring_params}
\begin{tabular}{lcc}
\toprule
 & \textbf{Biology} & \textbf{Chemistry} \\
\midrule
Knowledge aggregation & family MPool$_8$ & top-16 mean \\
Reasoning aggregation & gated MPool$_3$ & top-5 mean \\
Edge aggregation & target MPool$_8$ & top-16 weighted mean \\
Path aggregation & reliability + MPool$_8$ & top-16 mean \\
Directional Top-$k$ & 3 & 1 \\
Directional weight $\eta$ & 0.30 & 0.10 \\
\bottomrule
\end{tabular}
\end{table}
\section{Evaluation Protocol}
\label{app:evaluation_details}

\subsection{Scientific-Paper Evaluation Pool}
\label{app:paper_pool}

We evaluate SciGraphSelect on PeS2o~
\citep{peS2o} scientific-paper
documents spanning Biology, Chemistry, Physics, Mathematics, Environmental
Science, Geology, Geography, Agricultural and Food Sciences, and documents
with an unknown source-domain label. All documents are represented using
Qwen3-Embedding-0.6B. We construct one-vs-rest evaluations for Biology and
Chemistry, using the corresponding PeS2o source domain as positives and all
remaining source domains as negatives.

\begin{table}[t]
    \centering
    \small
    \caption{Composition of the full PeS2o paper evaluation.}
    \label{tab:pes2o_composition}
    \begin{tabular}{lrrr}
        \toprule
        Target domain & Positive & Negative & Prevalence \\
        \midrule
        Biology & $5{,}193{,}143$ & $9{,}919{,}169$ & $34.36\%$ \\
        Chemistry & $1{,}482{,}270$ & $13{,}630{,}042$ & $9.81\%$ \\
        \bottomrule
    \end{tabular}
\end{table}

These labels provide a source-controlled proxy for relevance rather than
document-level semantic ground truth; we therefore interpret the evaluation
as target-domain enrichment under source-proxy labels.

\subsection{Heterogeneous Hard-Negative Setting}
\label{app:hard_negative}

For each target domain, the hard-negative validation set contains 100,000
target-domain positives and 705,093 negatives. Negatives include adjacent
scientific papers, mathematical web data, mathematical books, mathematical
reasoning traces, general reasoning data, general web documents,
humanities/social-science/law data, and narrative/legal documents. Most strata
contain 100,000 examples; the mathematical-book stratum uses all 5,093
available examples. This set is used for directional-fusion selection, after
which the selected settings are fixed for PeS2o evaluation.

\subsection{Baselines and Selection Stages}
\label{app:baselines}

All embedding-based methods use Qwen3-Embedding-0.6B~\cite{zhang2025qwen3embedding} document
representations. \emph{Embedding similarity} uses the
maximum similarity to an extracted knowledge node. \emph{Concept-only} applies
the domain-specific knowledge aggregation. \emph{Relation co-activation} adds
reference-weighted relation co-support. \emph{Reasoning/path} adds reasoning
coverage and attested-path evidence. \emph{Structure-aware} denotes the
corresponding domain-specific graph score before directional fusion.
\emph{Directional alignment} evaluates the direction-only margin score, and
\emph{Final direction-aware} fuses percentile-normalized graph and
directional scores as described in Appendix~\ref{app:direction_fusion}.

For adjacent stages, top-set turnover compares the top-1\% selections.
\emph{Retained} denotes documents present in both sets; \emph{entered} and
\emph{exited} denote documents added to or removed from the new selection.
TP/FP use PeS2o source-domain proxy labels, $\Delta$TP is the net change in
target-domain documents at fixed budget, and turnover is the fraction of the
new top set not retained from the preceding stage.

\subsection{Evaluation Metrics}
\label{app:metrics}

For a scoring function $S$ and selection fraction $b$, let
$\mathcal{T}_b(S)$ denote the highest-scoring $\lceil bN\rceil$ documents.
\begin{equation}
\operatorname{P@}b
=
\frac{\sum_{i\in\mathcal{T}_b(S)}y_i}{|\mathcal{T}_b(S)|},
\end{equation}
where $y_i\in\{0,1\}$ is the source-proxy target label. We report P@1\%,
P@5\%, P@10\%, AUPRC, and AUROC. Stage-transition analyses additionally report
$\Delta$TP and turnover at the fixed top-1\% budget.

\subsection{Human Evaluation Protocol}
\label{app:human_eval_protocol}

The human evaluation complements PeS2o source-domain proxy labels with
blinded judgments of document relevance and attribution support.
The main text reports the aggregate results; here we describe the sampling
and annotation procedure.

\begin{figure}[ht]
    \centering
    \begin{subfigure}[t]{0.49\linewidth}
        \centering
        \includegraphics[width=\linewidth]{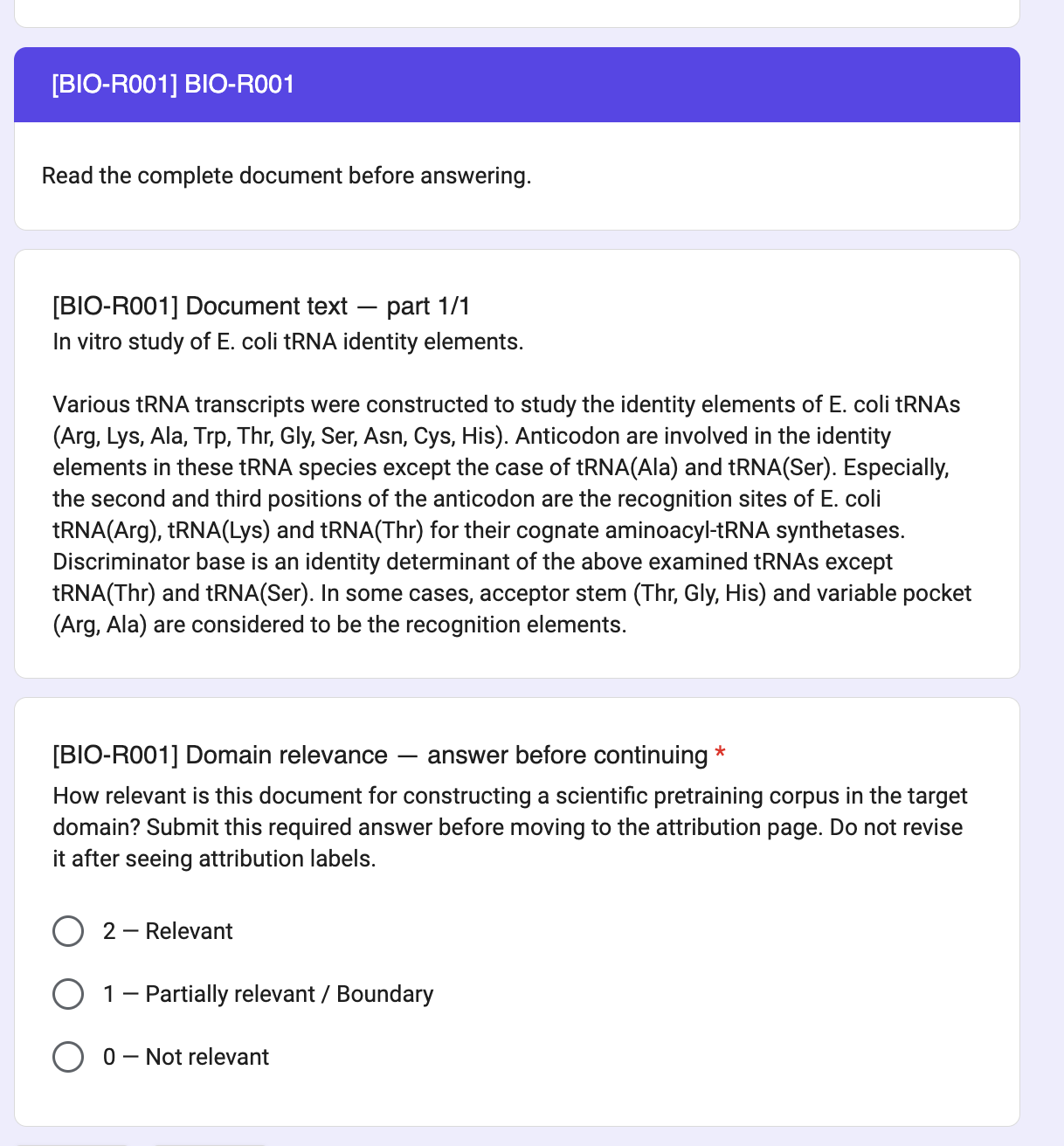}
        \caption{Document relevance judgment.}
    \end{subfigure}
    \hfill
    \begin{subfigure}[t]{0.49\linewidth}
        \centering
        \includegraphics[width=\linewidth]{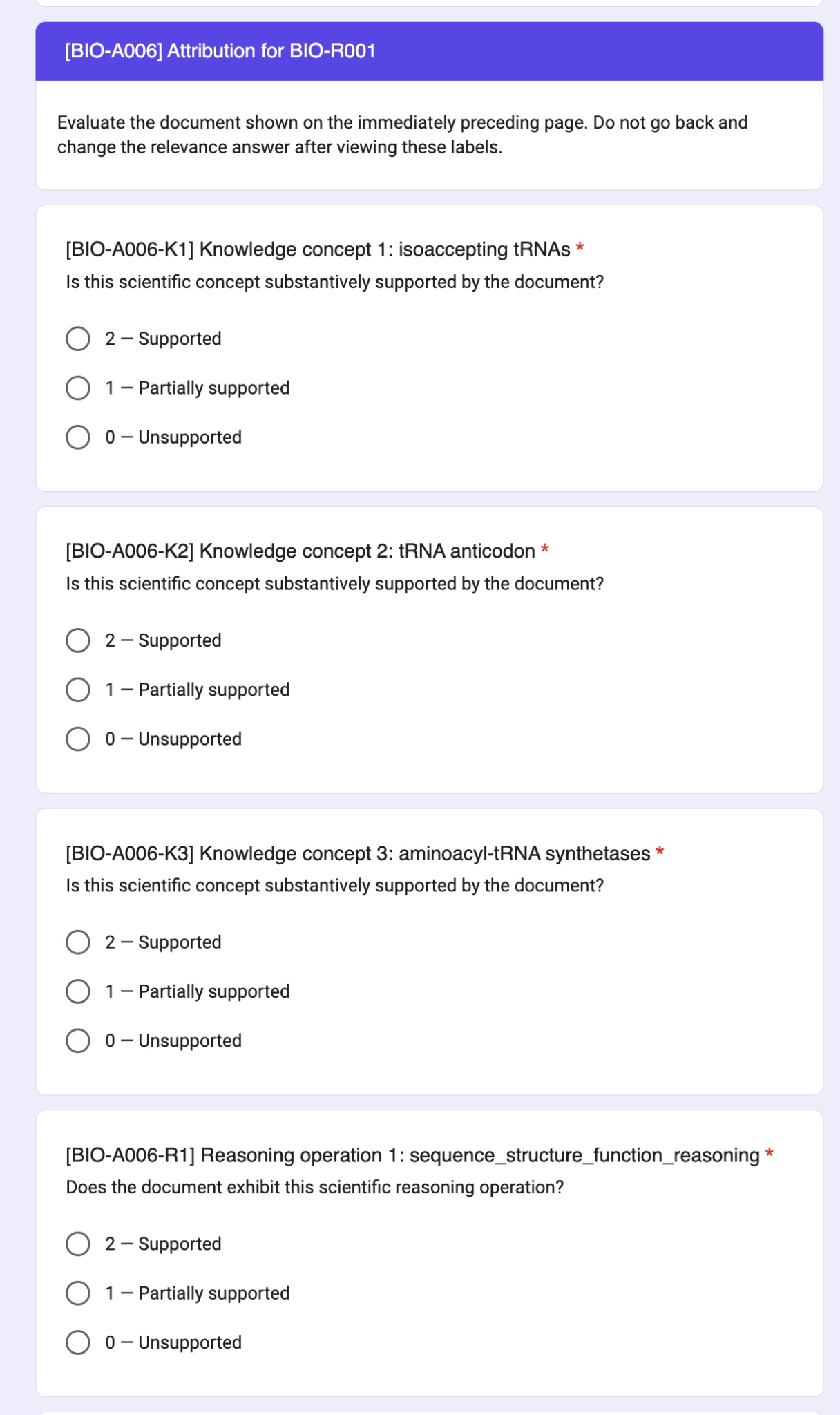}
        \caption{Attribution-support judgment.}
    \end{subfigure}
    \caption{
    \textbf{Blinded human-evaluation interface.}
    Annotators first read the candidate document and assess its relevance to
    the target scientific domain before seeing any SciGraphSelect attribution.
    They then evaluate the support for attributed knowledge concepts,
    reasoning operations, and relations. Relevance judgments are fixed before
    attribution labels are revealed.
    }
    \label{fig:human_annotation_interface}
\end{figure}

\paragraph{Selection relevance.}
For each domain, we sample 60 documents: 25 selected only by embedding
similarity, 25 selected only by SciGraphSelect, and 10 selected by both.
The unique subsets form the primary comparison, while the shared subset serves
as a blinded control.

\begin{table}[t]
\centering
\small
\caption{Human relevance sample per domain.}
\label{tab:human_relevance_sampling}
\begin{tabular}{lrrr}
\toprule
\textbf{Domain} & \textbf{Embedding-only} & \textbf{SciGraphSelect-only}
& \textbf{Both} \\
\midrule
Chemistry & 25 & 25 & 10 \\
Biology   & 25 & 25 & 10 \\
\bottomrule
\end{tabular}
\end{table}

As shown in Figure~\ref{fig:human_annotation_interface}(a), annotators first
read the full document and assign one of three relevance levels:
\emph{relevant}, \emph{partially relevant/boundary}, or \emph{not relevant}.
They are instructed to submit this judgment before proceeding to the
attribution page and not to revise it after seeing the attributed labels.
The selection method, PeS2o source-domain label, score, and rank are hidden
throughout annotation.

\paragraph{Attribution support.}
For each domain, we sample 30 SciGraphSelect-selected documents: 20 selected
only by SciGraphSelect and 10 shared with embedding similarity.

\begin{table}[t]
\centering
\small
\caption{Human attribution sample per domain.}
\label{tab:human_attribution_sampling}
\begin{tabular}{lrr}
\toprule
\textbf{Domain} & \textbf{SciGraphSelect-only} & \textbf{Both} \\
\midrule
Chemistry & 20 & 10 \\
Biology   & 20 & 10 \\
\bottomrule
\end{tabular}
\end{table}

Each document is presented with three attributed knowledge concepts, two
reasoning operations, and two relations.
As illustrated in Figure~\ref{fig:human_annotation_interface}(b), concept and
reasoning attributions are independently rated as \emph{supported},
\emph{partially supported}, or \emph{unsupported}.
For relation attributions, annotators judge whether both attributed elements
are supported and are used in a scientifically compatible relationship.
This evaluates relation support without requiring explicit directional
entailment of the reference relation.
Inter-annotator agreement is summarized with Cohen's $\kappa$.
\section{Additional Analyses}
\label{app:additional_results}

\subsection{UMAP View of Biology Selections and Local Graph Evidence}
\label{app:umap_bio}
\begin{figure*}[t]
    \centering
    \includegraphics[width=\textwidth]
    {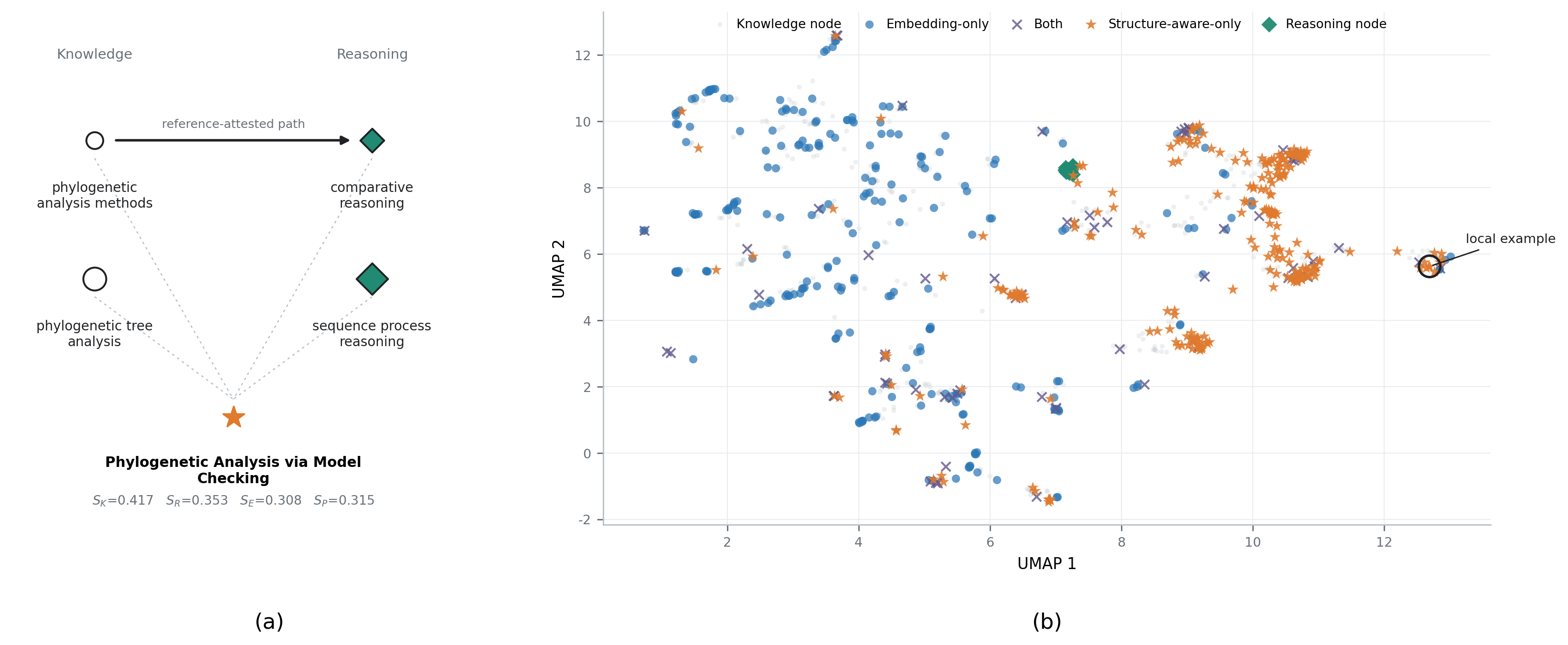}
    \caption{
    \textbf{Biology selection geometry and local graph evidence.}
    (a) Local evidence for a Biology document selected only by the
    structure-aware score.  Supporting knowledge and reasoning nodes are
    connected to the document, and the solid arrow marks a
    reference-attested knowledge-to-reasoning path.
    (b) Joint UMAP of documents selected only by embedding similarity,
    by both methods, or only by the structure-aware scorer, together with
    activated knowledge and reasoning nodes.  The circled point corresponds
    to the example in (a).  UMAP is used only for visualization; document
    selection and graph matching are performed in the original embedding
    space.
    }
    \label{fig:bio_selection_umap}
\end{figure*}

Figure~\ref{fig:bio_selection_umap} provides a Biology counterpart to the
Chemistry visualization in the main text.  In the joint projection,
embedding-only and structure-aware-only documents exhibit different
concentration patterns, while shared selections are interspersed across the
projected space.  The highlighted structure-aware-only document is supported
by phylogenetic-analysis knowledge and multiple reasoning nodes, including a
reference-attested path from \emph{phylogenetic analysis methods} to
\emph{comparative reasoning}.  This example illustrates how evidence
distributed across multiple graph elements can support a document that is not
retained by maximum embedding similarity alone.  As throughout
SciGraphSelect, the displayed path denotes reference-attested structure and
the document provides endpoint co-support; the visualization does not imply
that the candidate text entails the relation.

\subsection{Topology Placebo Tests}
\label{app:topology}

We examine whether the exact arrangement of reference-attested relations and
paths contributes selection signal beyond the node inventory and matched graph
statistics. For the edge control, relation targets are rewired within each
edge type while preserving node degrees and edge weights. For the path
control, nodes are shuffled across paths of the same type and length while
preserving their position-wise composition.

\begin{table}[t]
    \centering
    \small
    \caption{
    Topology-placebo results. Values denote the AUPRC of the true reference
    topology minus that of the corresponding placebo.
    }
    \label{tab:topology_appendix}
    \begin{tabular}{llc}
        \toprule
        \textbf{Domain} & \textbf{Comparison} & $\Delta$AUPRC \\
        \midrule
        Biology
        & True topology -- rewired edges
        & $+0.00361$ \\
        Biology
        & True topology -- shuffled paths
        & $+0.00134$ \\
        Chemistry
        & True topology -- rewired edges
        & $+0.00050$ \\
        Chemistry
        & True topology -- shuffled paths
        & $+0.00014$ \\
        \bottomrule
    \end{tabular}
\end{table}

The true topology yields higher AUPRC than both matched controls in Bio and
Chemistry. The improvement is larger for edge rewiring than for path
shuffling in both domains, suggesting that reference-attested connectivity
provides a small amount of additional selection signal beyond the node
inventory. Overall, the results support using the extracted relations and
paths as complementary structural evidence in SciGraphSelect.

Relation directionality is evaluated separately through forward--reverse
alignment, since reversing an edge does not change relation co-support.

\subsection{Counterfactual Evidence Removal}
\label{sec:counterfactual-removal}

We test whether the sentences identified by SciGraphSelect are relevant to its
scoring decisions under a deletion intervention.  For each document, we remove
the sentence most similar to the highest-scoring eligible knowledge or reasoning
node.  As a shared control, we remove a non-attributed sentence
from the same document whose word count is closest to that of the
concept-attributed sentence.  We select the 50 highest-scoring proxy-positive
documents in each domain.

Each deletion variant is re-embedded, and its component and graph scores
are recomputed using the exact graph scorer.  We then recompute the final
direction-aware score against the fixed validation-pool ECDF.  We define
\[
\Delta S = S(x)-S(x\setminus e),
\]
such that $\Delta S>0$ denotes a score decrease after deletion.  For each
attribution type $t\in\{K,R\}$, selectivity is defined as
\[
\operatorname{Sel}_t
=
\overline{\Delta S_{\mathrm{final}}^{(t)}}
-
\overline{\Delta S_{\mathrm{final}}^{(\mathrm{ctrl})}}.
\]
\begin{table*}[t]
\centering
\small
\caption{Counterfactual sentence removal under the final domain scorers.
Selectivity is the difference in final-score drop between attributed-sentence
deletion and the shared Shared control deletion.}
\label{tab:counterfactual-removal}
\begin{tabular}{lrrrrrr}
\toprule
\textbf{Removal}
& \textbf{$n$}
& \textbf{$\Delta S_K$}
& \textbf{$\Delta S_R$}
& \textbf{$\Delta S_{\mathrm{graph}}$}
& \textbf{$\Delta S_{\mathrm{final}}$}
& \textbf{Selectivity} \\
\midrule

\multicolumn{7}{l}{\textbf{Chemistry}} \\
\quad Shared control
& 50
& $-0.03162$
& $-0.04094$
& $-0.03460$
& $-0.000152$
& -- \\

\quad Attributed concept
& 50
& $+0.01000$
& $-0.03190$
& $+0.01859$
& $+0.000046$
& $+0.000198$ \\

\quad Attributed reasoning
& 50
& $+0.01164$
& $-0.02518$
& $+0.01633$
& $+0.000044$
& $+0.000196$ \\

\midrule

\multicolumn{7}{l}{\textbf{Biology}} \\
\quad Shared control
& 50
& $-0.00428$
& $-0.03216$
& $-0.00630$
& $+0.000287$
& -- \\

\quad Attributed concept
& 50
& $+0.03214$
& $-0.03044$
& $+0.02986$
& $+0.004237$
& \textbf{$+0.003950$} \\

\quad Attributed reasoning
& 50
& $+0.01242$
& $-0.01679$
& $+0.01112$
& $+0.002426$
& \textbf{$+0.002139$} \\

\bottomrule
\end{tabular}
\end{table*}
Biology exhibits selective counterfactual sensitivity: deleting attributed
concept or reasoning evidence reduces the structure-aware and final scores more than
the Shared control deletion, with a stronger effect for concept
evidence.  Because reasoning-targeted deletion does not consistently reduce
the reasoning component itself, however, the result supports aggregate score
relevance rather than component-specific faithfulness.

Chemistry exhibits the same relative effect at the structure-aware stage, before ECDF normalization, but it is
largely attenuated at the final ECDF-calibrated score.  The Chemistry result
therefore provides limited evidence for final-score faithfulness.  Since
sentence deletion also perturbs the whole-document representation, positional
context, and truncation boundary, we interpret the experiment as measuring
selective counterfactual sensitivity rather than the isolated causal
contribution of an attributed semantic unit.

\section{Qualitative Examples and Failure Cases}
\label{sec:qualitative-examples}

\subsection{Successful Selections}

We inspect representative high-scoring candidates from the final
direction-aware scorers. To make the selection rationale directly auditable,
we report excerpts from the actual scored documents together with their
component scores and activated graph evidence.

\begin{successbox}{Biology: Knowledge + reasoning
(\texttt{hardneg::biology::00020553})}
\scriptsize

\textbf{Candidate excerpt.}
\emph{``GoFish finds genes with combinations of Gene Ontology attributes.
GoFish is a Java application that allows users to search for gene products
with particular gene ontology (GO) attributes, or combinations of attributes.
GoFish ranks gene products by the degree to which they satisfy a Boolean
query.''}

\vspace{0.35em}
\textbf{Scores.}
$(S_K,S_R,S_E,S_P)=(.3801,.3734,.0006,.1290)$;
graph $=.4027$, final $=.9948$.

\vspace{0.35em}
\textbf{Activated graph evidence.}
\emph{Gene Ontology database}, together with
sequence--structure--function and phylogenetic reasoning.

\vspace{0.35em}
\textbf{Interpretation.}
The document directly concerns GO-based gene-product retrieval.
Its selection is well explained by knowledge and reasoning support even though
relation co-support is weak, illustrating that SciGraphSelect does not require
every useful document to instantiate a strong relation or path.
\end{successbox}

\begin{successbox}{Biology: Knowledge-dominant selection
(\texttt{hardneg::biology::00012265})}
\scriptsize

\textbf{Candidate excerpt.}
\emph{``The aim of the present work is to characterize strains that can be
used for broad-spectrum pest control in agriculture. Twenty strains of
B. thuringiensis were isolated from Bulgarian soil habitats. The strains were
screened for genes encoding 12 different crystal (Cry) endotoxins by PCR with
specific primer pairs. Seven of the isolates contained cry genes in their
genomes.''}

\vspace{0.35em}
\textbf{Scores.}
$(S_K,S_R,S_E,S_P)=(.4288,.0264,0,0)$;
graph $=.4300$, final $=.9944$.

\vspace{0.35em}
\textbf{Activated graph evidence.}
\emph{Bt toxin genes}, \emph{Bt gene},
\emph{B. thuringiensis}, and \emph{Bt toxin}.

\vspace{0.35em}
\textbf{Interpretation.}
Specific biological concepts are recovered without relation or path
activation. This case shows that standalone knowledge coverage remains a
valid selection signal rather than forcing every scientific document into a
reasoning-chain representation.
\end{successbox}

\begin{successbox}{Chemistry: Structured $K\!\rightarrow\!R$ selection
(\texttt{hardneg::chemistry::00043663})}
\scriptsize

\textbf{Candidate excerpt.}
\emph{``The reactivity of a reported hyper-coordinated alkene
(Rea-E; Rea = reactant; E = group 14 element) featuring a central
E[double bond]E moiety was theoretically analyzed using DFT
(density functional theory) and the EDA-NOCV
(energy decomposition analysis-natural orbitals for chemical valence)
method. M06-2X/def2-SVP and B3LYP-D3/def2-SVP results demonstrate that five
Rea-E molecules have an energy minimum as their structures have no imaginary
frequency.''}

\vspace{0.35em}
\textbf{Scores.}
$(S_K,S_R,S_E,S_P)=(.3667,.3603,.3251,.3478)$;
graph $=.3546$, final $=.9998$.

\vspace{0.35em}
\textbf{Activated graph evidence.}
\emph{Enone/alkene reactivity}, orbital reasoning,
reaction-mechanism reasoning, and the reference-supported association
\emph{enone reactivity $\rightarrow$ reaction-mechanism reasoning}.

\vspace{0.35em}
\textbf{Interpretation.}
Knowledge, reasoning, relation co-support, and path evidence all correspond to
the document's computational-mechanistic analysis, illustrating structured
selection beyond a single nearest-concept match.
\end{successbox}

\begin{successbox}{Chemistry: Mechanistic-path selection
(\texttt{hardneg::chemistry::00024847})}
\scriptsize

\textbf{Candidate excerpt.}
\emph{``In the presence of multiple electrophiles, the reaction sequence is a
critical mechanistic problem. Here, we report a theoretical study on the
mechanism of phosphine-catalyzed [8+2] cycloaddition of heptafulvenes and
allenoate. DFT calculations showed that electrophilicity is the barrier for
nucleophilic attack, while it fails in the prediction of priority.''}

\vspace{0.35em}
\textbf{Scores.}
$(S_K,S_R,S_E,S_P)=(.4043,.2982,.3358,.3529)$;
graph $=.3670$, final $=.9998$.

\vspace{0.35em}
\textbf{Activated graph evidence.}
\emph{Phosphine-catalyzed reactions},
\emph{electrophilic addition}, and
\emph{electrophilic addition $\rightarrow$ reaction-mechanism reasoning}.

\vspace{0.35em}
\textbf{Interpretation.}
The candidate explicitly discusses both chemical reactivity and mechanistic
analysis, so its structural evidence is supported by substantive Chemistry
content rather than generic reasoning language.
\end{successbox}

\paragraph{Recovery from mixed-source data.}
SciGraphSelect also identifies scientifically relevant documents whose source
labels place them outside the target domain.

\begin{successbox}{Biology: Relevant content hidden by source label
(\texttt{hardneg::biology::00478056})}
\scriptsize

\textbf{Candidate excerpt.}
\emph{``In the context of neural plasticity, which of the following best
describes the process of long-term potentiation (LTP)? [\ldots]
D: LTP enhances the strength of synaptic connections, leading to improved
neural communication.''}

\vspace{0.35em}
\textbf{Activated evidence.}
The document matches \emph{Long-term potentiation} and
\emph{long-term depression}, with $S_K=.4922$ and final score $.9984$.

\vspace{0.35em}
\textbf{Interpretation.}
Although its source label places it among general-reasoning negatives, the
document contains direct neuroscience content. This is therefore a
source-proxy false positive rather than a semantic selection error.
\end{successbox}

\begin{successbox}{Chemistry: Relevant content hidden by source label
(\texttt{hardneg::chemistry::00447865})}
\scriptsize

\textbf{Candidate excerpt.}
\emph{``Which of the following analytical techniques is most suitable for the
identification of organic compounds based on their molecular structure?
[\ldots] E: Nuclear Magnetic Resonance (NMR).
F: Infrared Spectroscopy (IR).''}

\vspace{0.35em}
\textbf{Scores.}
$(S_K,S_R,S_E,S_P)=(.4284,.3196,.2642,.2642)$;
final $=.9998$.

\vspace{0.35em}
\textbf{Activated graph evidence.}
IR/NMR-based molecular identification and
spectroscopic structure elucidation.

\vspace{0.35em}
\textbf{Interpretation.}
The candidate contains substantive analytical Chemistry despite its
non-target source label, again illustrating the limitations of source-domain
labels as exact semantic ground truth.
\end{successbox}

\subsection{Failure Cases}

The following examples are genuine high-scoring failures: the source text does
not contain substantive target-domain content, but generic concepts or
reasoning operations activate parts of the reference graph.

\begin{failurebox}{Biology failure: Generic statistical concept
(\texttt{hardneg::biology::00292356})}
\scriptsize

\textbf{Candidate excerpt.}
\emph{``Have you ever heard of the word `average' in math? Have you ever tried
to figure out the average or mean of a set of numbers? Tania and Alex are
planning for next year's garden. Tania has decided to complete a harvesting
review of carrots. She wants to use the number of carrots picked each week to
make conclusions about carrot growth.''}

\vspace{0.35em}
\textbf{Scores.}
$(S_K,S_R,S_E,S_P)=(.3404,.1339,0,0)$;
graph $=.3475$, final $=.9827$.

\vspace{0.35em}
\textbf{Activated graph evidence.}
\emph{Mean} and \emph{population mean}, together with quantitative and
statistical reasoning.

\vspace{0.35em}
\textbf{Failure mechanism.}
The document contains no substantive Biology. Generic statistical concepts
entered the graph through quantitative biological references but retain
surface forms broad enough to match elementary mathematics.
This case motivates stronger specificity constraints on generic knowledge
nodes.
\end{failurebox}

\begin{failurebox}{Chemistry failure: Narrative reasoning collision
(\texttt{hardneg::chemistry::00767616})}
\scriptsize

\textbf{Candidate excerpt.}
\emph{``But, (herein, me beleeve) Thy studie for meere-praise, will thee
deceive: And, if thy Vertues, be, but onely, those For which the vulgar Fame,
her Trumpet blowes, Thy Fame's a blast; Thy Vertues, Vices be; Thy Studie's
vaine; and, shame will follow thee.''}

\vspace{0.35em}
\textbf{Scores.}
$(S_K,S_R,S_E,S_P)=(.1521,.3515,.2777,.2966)$;
graph $=.2151$, final $=.9867$.

\vspace{0.35em}
\textbf{Activated graph evidence.}
Weak knowledge collisions with laboratory safety and measurement combine with
generic operations such as concept/rule application, factual recall, evidence
synthesis, and $R\!\rightarrow\!R$ path activation.

\vspace{0.35em}
\textbf{Failure mechanism.}
Reusable reasoning nodes can activate even when domain-specific Chemistry
content is absent. The example motivates stronger anchoring of reasoning and
path evidence to domain-specific knowledge.
\end{failurebox}

\begin{failurebox}{Chemistry failure: Generic quantitative reasoning
(\texttt{hardneg::chemistry::00264091})}
\scriptsize

\textbf{Candidate excerpt.}
\emph{``The expression $(x^2-y^2)/(x-y)$ can be simplified by recognizing it
as a difference of squares:
$x^2-y^2=(x+y)(x-y)$. Thus, the expression simplifies to $x+y$, provided
$x\ne y$.''}

\vspace{0.35em}
\textbf{Scores.}
$(S_K,S_R,S_E,S_P)=(.2848,.5357,.4000,.4273)$;
graph $=.3668$, final $=.9961$.

\vspace{0.35em}
\textbf{Activated graph evidence.}
\emph{Linear equations}, quantitative calculation, equilibrium reasoning,
thermodynamic reasoning, and dense reasoning-path evidence.

\vspace{0.35em}
\textbf{Failure mechanism.}
The generic mathematical procedure strongly matches reusable reasoning
operations despite lacking Chemistry content. This failure motivates requiring
reasoning/path evidence to remain sufficiently anchored to domain-specific
knowledge and $K\!\rightarrow\!R$ structure.
\end{failurebox}



\end{document}